\documentclass[letterpaper, 10 pt, conference]{ieeeconf}  

\IEEEoverridecommandlockouts                              
\usepackage{graphics} 
\usepackage{epsfig} 
\usepackage{mathptmx} 
\usepackage{times} 
\usepackage{amsmath} 
\usepackage{amssymb}  
\usepackage{amsmath}
\usepackage{tabularx, booktabs, multirow, siunitx}

\title{\LARGE \bf
RL-Based Coverage Path Planning for Deformable Objects on 3D Surfaces
}
\author{Yuhang Zhang$^{1}$, Jinming Ma$^{2}$ and Feng Wu$^{1*}$%
\thanks{This work was supported in part by Major Research Plan of the National Natural Science Foundation of China (92048301), Anhui Provincial Major Research and Development Plan (202004H07020008), and Anhui Provincial Natural Science Foundation (2208085MF172).}%
\thanks{$^{1}$Yuhang Zhang and Feng Wu are both with the School of Computer Science and Technology, University of Science and Technology of China, Hefei, Anhui, China. {\tt\small zyh23011146@mail.ustc.edu.cn, wufeng02@ustc.edu.cn}, $^{2}$Jinming Ma is with Xiaomi Robotics Lab, Beijing, China. {\tt\small majinming3@xiaomi.com}}%
\thanks{$^{*}$ Feng Wu is the corresponding author.}%
}

\begin{document}

\maketitle
\thispagestyle{empty}
\pagestyle{empty}

\begin{abstract}
Currently, manipulation tasks for deformable objects often focus on activities like folding clothes, handling ropes, and manipulating bags. However, research on contact-rich tasks involving deformable objects remains relatively underdeveloped. 
When humans use cloth or sponges to wipe surfaces, they rely on both vision and tactile feedback. Yet, current algorithms still face challenges with issues like occlusion, while research on tactile perception for manipulation is still evolving. Tasks such as covering surfaces with deformable objects demand not only perception but also precise robotic manipulation.
To address this, we propose a method that leverages efficient and accessible simulators for task execution. Specifically, we train a reinforcement learning agent in a simulator to manipulate deformable objects for surface wiping tasks.
We simplify the state representation of object surfaces using harmonic UV mapping, process contact feedback from the simulator on 2D feature maps, and use scaled grouped convolutions (SGCNN) to extract features efficiently. The agent then outputs actions in a reduced-dimensional action space to generate coverage paths. Experiments demonstrate that our method outperforms previous approaches in key metrics, including total path length and coverage area. We deploy these paths on a Kinova Gen3 manipulator to perform wiping experiments on the back of a torso model, validating the feasibility of our approach. 
\end{abstract}

\section{INTRODUCTION}

In the Complete Coverage Path Planning (CPP) problem, a robot is required to find a path that covers the entire free space of a confined area. This classical problem has broad practical applications in the field of automation, with typical scenarios including lawn mowing \cite{hameed2017coverage}, autonomous cleaning \cite{viet2013ba}, and agriculture \cite{hameed2014intelligent}. Traditional CPP algorithms often make rigid assumptions about the environment, simplifying the operational space into a two-dimensional plane or a known static three-dimensional model. Based on this assumption, systematic solutions such as grid-based decomposition \cite{wieser2014autonomous} and cellular decomposition \cite{wang2015coverage} have been developed. These methods have demonstrated effectiveness in achieving coverage and operational efficiency in static environments \cite{hou2025spiral}.

However, as robotics advances, the limitations of traditional methods become increasingly exposed in complex, unstructured environments. One significant challenge is achieving high-quality complete coverage on target surfaces using deformable objects (e.g., sponges or soft cloths). 
Unlike the manipulation of rigid objects, tasks involving deformable materials require robotic systems to do more than simply plan coverage paths over a surface. These systems must also be able to dynamically adapt to the changes that occur during surface contact, including object deformation, stretching, and compression. 
For example, medical rehabilitation procedures, such as disinfection, dressing, surgery, or massage performed on the human torso or limbs \cite{thach2023deformernet}, often involve deformable objects that must conform to the complex geometry of the human body.
Similarly, tasks like coverage path planning with a sponge on 3D surfaces \cite{le2023sponge} introduce additional challenges, as the sponge’s physical properties—such as flexibility, absorbency, and friction—change based on the surface contours and the force applied. In such scenarios, achieving efficient coverage requires not only precise planning but also real-time sensory feedback and adjustment to accommodate the dynamic interaction between the deformable object and surface.

In summary, complete coverage path planning for interactions between deformable objects and 3D surfaces still faces significant open challenges, including accurately acquiring real-time contact information during manipulation, as well as optimizing the total path length and coverage area while accounting for object deformation. Addressing these challenges will advance the effectiveness and reliability of robotic systems in automation scenarios.

\begin{figure}[!t]
    \centering
    \includegraphics[width=0.95\linewidth]{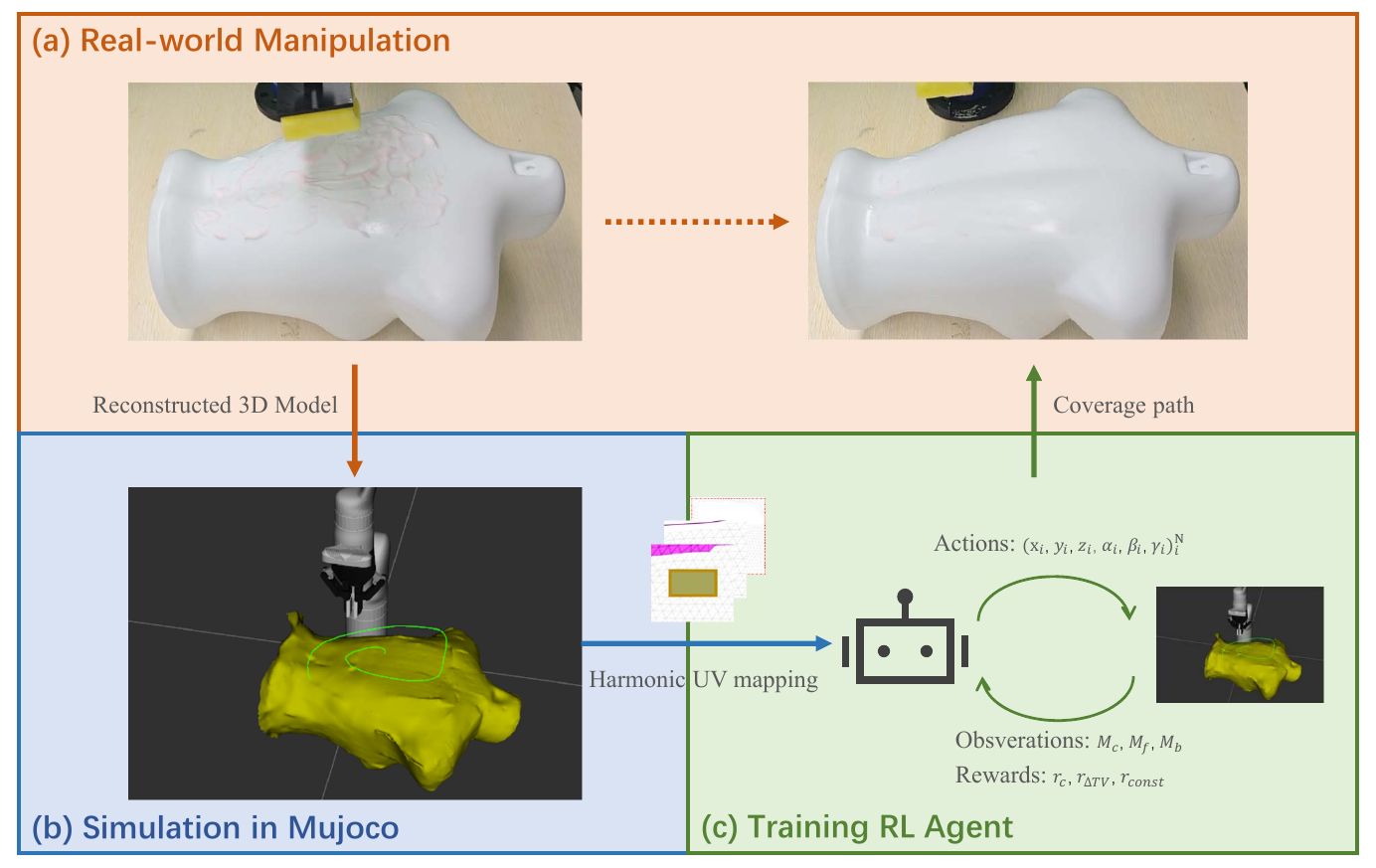}
    \caption{\textbf{Overview of Framework.} 
    We propose a framework to solve the problem of using deformable objects to cover 3D surfaces. 1) A 3D model of the target object is reconstructed, and a wiping task environment is created within Mujoco. 2) By employing harmonic UV mapping, we simplify the state representation and action space. 3) The reinforcement learning algorithm outputs an efficient coverage path to cover the surface.}
    \label{fig:placeholder}
\end{figure}

To address the coverage path planning task for deformable objects on 3D surfaces, we propose a novel pipeline that integrates \emph{force feedback simulation, dimensionality reduction via UV mapping, and a reinforcement learning agent equipped with SGCNN as the feature extractor}.
In this pipeline, we first acquire low-cost force feedback between deformable objects and surfaces within a simulation environment, enabling real-time interaction data. Then, we apply harmonic UV mapping to reduce the dimensionality of both the state and action spaces, simplifying the problem for reinforcement learning. By employing SGCNN \cite{10.5555/3692070.3692973} as the feature extractor, the RL agent outputs actions that ultimately generate paths for deformable objects to efficiently cover 3D surfaces. 

Our contributions are as follows:

\begin{enumerate}
    \item \textbf{Simplified State and Action Spaces:} We introduce harmonic UV mapping to project 3D object surfaces onto a 2D plane, significantly simplifying the state and action spaces for the reinforcement learning algorithm, thus enhancing computational efficiency and convergence speed.

    \item \textbf{Efficient Pipeline for Surface Coverage:} We propose an innovative processing pipeline tailored to deformable object surface coverage tasks. The method demonstrates substantial improvements in key performance metrics, including wiping path length and coverage area, compared to existing solutions.
\end{enumerate}

Finally, we validate the effectiveness and feasibility of our proposed method through simulation and real-world experiments. In these experiments, the system successfully adapts to deformable objects and a variety of 3D surface types, confirming the practical applicability of our approach.

\section{RELATED WORK}
This section reviews related work in three aspects: traditional coverage path algorithms, learning-based planning and control algorithms, and modeling and manipulation of deformable objects.

\subsection{Traditional Coverage Path Algorithms}

Previous researchers have conducted extensive studies on traditional methods for surface path planning. These methods require parameterization of the object surface and sampling of path points in the parameter domain based on rules. For example, Sheng \textit{et al.} \cite{sheng2005tool} proposed a rule-based method that first partitions the surface of a 3D model into a series of blocks and then determines the motion pattern and scanning direction of the actuator for each block, choosing between zigzag and spiral motion patterns. Chen \textit{et al.} \cite{chen2017trajectory} introduced a Bézier curve-based trajectory optimization method for robots spraying object surfaces, which samples path points in the parameter domain of the Bézier surface and optimizes them based on a spraying model. McGovern \textit{et al.} \cite{mcgovern2023general} considered the feasibility of executing parameter surface-based paths planned by robotic arms on object surfaces. These methods require models of both the actuator and the object surface and do not account for potential deformations of the actuator or the object surface.

\subsection{Learning-Based Planning and Control Algorithms}
Due to the complex and variable states of deformable objects, it is difficult to obtain their state information in real-world environments. Simulation software such as Mujoco \cite{todorov2012mujoco} and SOFA \cite{faure2012sofa} enables humans to collect data at a lower cost. Since reinforcement learning algorithms require large amounts of data for complex tasks, training intelligent agents in simulation environments is a favorable approach. Matas \textit{et al.}\cite{matas2018sim} employed the DDPGfD algorithm \cite{vecerik2017leveraging} for cloth manipulation tasks, training agents in a domain-randomized simulation environment and successfully transferring them to the real world. Jangir \textit{et al.} \cite{jangir2020dynamic} applied the DDPG algorithm \cite{lillicrap2015continuous} to train cloth folding in the SOFA simulation environment. However, obtaining cloth state information in the real world remains challenging. Hietala \textit{et al.} \cite{hietala2022learning} built on this work by incorporating visual images as input for the agent, successfully transferring the policy to the real world. Reinforcement learning algorithms have also been effectively applied in CPP tasks. Theile \textit{et al.} \cite{theile2020uav} utilized multi-channel images as input for UAV tasks with flight range constraints. Kiemel \textit{et al.} \cite{kiemel2019paintrl} implemented reinforcement learning for a robotic arm spraying task on car door surfaces, representing the spraying state by partitioning the surface into blocks. Jonnarth \textit{et al.} \cite{10.5555/3692070.3692973} adopted the SAC \cite{haarnoja2018soft} to enable agents to plan complete coverage paths in unknown environments. These studies operated in a two-dimensional action space. To enable robots to perform coverage operations on 3D object surfaces, we map both the state and action spaces to a 2D plane and process state information using SGCNN \cite{10.5555/3692070.3692973}. This approach not only facilitates efficient feature learning but also simplifies the dimensionality of the action space.

\subsection{Modeling and Manipulation of Deformable Objects}

Deformable objects undergo deformation when subjected to forces, and dynamic models predict this process. Becker \textit{et al.} \cite{becker2007robust} performed finite element modeling of deformable objects while identifying model parameters. Vazquez \textit{et al.} \cite{vazquez2024prediction} proposed an LSTM network that takes the original contour and applied forces as input to predict object deformation. Valencia \textit{et al.} \cite{valencia2020combining} and Liu \textit{et al.} \cite{liu2023softgpt} employed graph neural networks for this process. Chen et al. \cite{chen2021ab} and Chen \textit{et al.} \cite{chen2024differentiable} utilized differentiable particle dynamics models to infer robot actions. Yu \textit{et al.} \cite{yu2022shape} trained a neural network to serve as the Jacobian matrix between robotic arm joints and key points of a rope, thereby establishing control equations. Le \textit{et al.} \cite{le2023sponge} leveraged a Pointnet network \cite{qi2017pointnet} in a simulation environment to predict contact points between sponge and object. After developing this network, they sampled a series of path points from unpredicted contact points and finally generated the shortest path using a traveling salesman solver.

\section{METHOD}

\begin{figure*}[!t]
    \centering
    \includegraphics[width=0.95\linewidth]{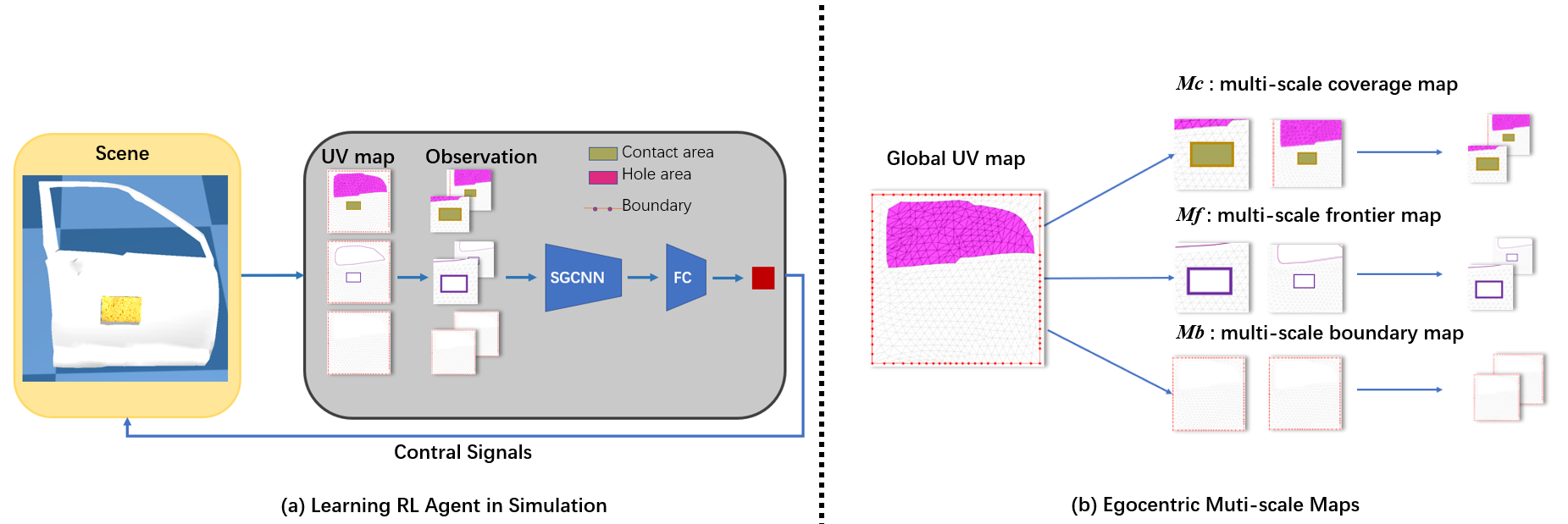}
    \caption{The proposed method first maps the target wiping area to the UV coordinate system via UV mapping. To represent the state more efficiently, we construct an agent-centric map representation. The coverage map, border map, and frontier map are translated, rotated, and scaled relative to the agent’s perspective. These multi-scale maps are then processed by an SGCNN \cite{10.5555/3692070.3692973} module, and control signals are finally output through fully-connected (FC) layers. Here shows two scales with a factor of 2. The agent is positioned at the center of the scaled maps. The scaled maps are discretized to the same resolution of $64\times64$ pixels. }
    \label{fig:placeholder}
\end{figure*}

In this section, we introduce an RL-based algorithm for the coverage path planning task for deformable objects on 3D surfaces. 
First, we propose to leverage a simulator to obtain contact information between deformable objects manipulated by a robot and target surfaces. Although acquiring such data in the real world is difficult and costly, simulation environments offer a low-cost alternative for collecting this information.
Then, we constrain the action space of the sponge to the object surface using parametric mapping. This restriction reduces the degree of freedom of the task, simplifying the learning process. 
Furthermore, to enhance the efficiency of the RL policy, we reduce the dimensionality of the contact state on the object surface from 3D to 2D. 
Finally, a 2D feature extractor is employed to capture the essential state features, allowing the RL agent to focus on the most relevant information for manipulation policy learning.

\subsection{Problem Formulation} 

Our task involves manipulating a deformable object to achieve coverage on a given target surface of interest, where coverage is defined as the contact between the deformable object and the surface. In the context of a sponge wiping task, the target surface can be modeled via 3D reconstruction, with the surface of the 3D model represented as a triangle mesh. The region of interest is extracted from the model mesh as the target wiping surface.

We formulate this surface coverage path planning task as a Markov Decision Process (MDP). The reinforcement learning policy network observes the current state and outputs an action for the current time step. The environment then returns a reward and transitions to the next state.

\subsection{UV Mapping}

We represent the human body model as a 3D mesh $\mathcal{M}$, which consists of a set of mesh vertices $\mathcal{V} = \{v_1,\ldots,v_n\}$and a collection of triangular faces $\mathcal{F}$. We extract the surface of interest and, assuming it is homeomorphic to a disk, map it onto a two-dimensional plane bounded by a convex hull. Parametric mapping methods can be broadly categorized into two types: fixed-boundary methods and free-boundary methods. To facilitate easier processing of states by the reinforcement learning agent and to constrain the motion range of the end effector, we adopt a fixed-boundary approach. Let the mesh surface of interest $\mathcal{F}_1$ contain $\mathcal{N}$ vertices, including $\mathcal{K}$ boundary points. We apply chordal parameterization to the boundary vertices $v_i$ as follows:
\begin{equation}
t_i = \frac{\sum_{j=0}^{i-1} \|\mathbf{v}_{j+1} - \mathbf{v}_j\|}{\sum_{j=0}^{K-1} \|\mathbf{v}_{j+1} - \mathbf{v}_j\|}, \quad i = 1, \ldots, K-1
\label{eq:chordal_parameterization}
\end{equation}
where $t_0=0$.

For the square domain $[-1,1]^2$, the parameterization results for the boundary vertices are:
\begin{equation}
(x_i, y_i) = 
\begin{cases}
(8t_i-1, -1) & t_i \in [0, 0.25) \\
(1, 8t_i - 3) & t_i \in [0.25, 0.5) \\
(5 - 8t_i, 1) & t_i \in [0.5, 0.75) \\
(-1, 7 - 8t_i) & t_i \in [0.75, 1]
\end{cases}
\label{eq:square_mapping}
\end{equation}

For the unit circle domain, the central angle corresponding to the arc between the starting boundary point and boundary point $v_i$is $2\pi t_i$. The parameterization results for the boundary vertices are:
\begin{equation}
(x_i, y_i) = (\cos(2\pi t_i), \sin(2\pi t_i)), \quad t_i \in [0, 1]
\label{eq:circle_mapping}
\end{equation}

\begin{figure*}[!t] 
    \centering
    \includegraphics[width=0.9\linewidth]{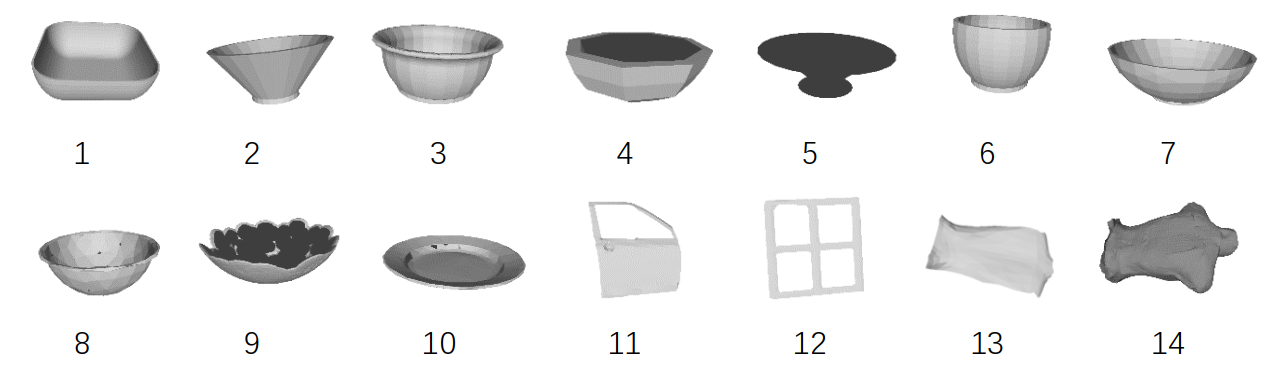}
    \caption{Objects 1-10 are sourced from the SPONGE dataset \cite{le2023sponge} and are used for quantitative analysis. The subsequent objects are employed for experiments on more complex geometries, including car doors, windows, and human body models.}
    \label{fig:objects}
\end{figure*}

Since the boundary conditions are known, we perform a harmonic mapping on the mesh to obtain the parametric coordinates of the interior vertices and the corresponding triangular faces $\mathcal{F}_{uv}$in the mapped domain. There exists a one-to-one correspondence between $\mathcal{F}$and $\mathcal{F}_{uv}$. Based on the properties of harmonic mapping, we can easily compute the correspondence between points on the parametric domain faces and points on the model surface using weight calculations. The parameterized surface $\mathbf{S}(u, v)$is defined as:
\begin{equation}
   \mathbf{S}(u, v) = 
    \begin{pmatrix}
    x(u, v) \\
    y(u, v) \\
    z(u, v)
    \end{pmatrix} 
\end{equation}
where $x(u, v)$, $y(u, v)$, and $z(u, v)$ are the coordinate components of the surface at the parameter $(u, v)$.

\subsection{Observation Space}
In surface coverage tasks, effective representing the state is crucial. If the object surface is considered as a discrete point set $\mathcal{V}$ containing $N$ vertices, then simply representing whether each point is covered would require a state space of size $2^N\times N$, which undoubtedly poses significant challenges for the convergence of reinforcement learning agents. Therefore, to preserve the topological structure of the mesh surface while enabling efficient state representation, we propose representing the state in a two-dimensional space and using convolutional neural networks to extract features effectively.

After mapping the object surface to a square or unit circular domain via UV mapping, we obtain the corresponding domain in the UV coordinate system through translation and scaling. Similar to the multi-scale map representation used by Jonnarth \textit{et al.} \cite{10.5555/3692070.3692973} for agent exploration tasks in 2D maps, we also adopt a multi-scale map approach. 

Our map consists of three layers: a \textbf{coverage map ($M_c$)}, a \textbf{border map ($M_b$)}, and a \textbf{frontier map ($M_f$)}. 
The \textbf{coverage map $M_c$} is a binary matrix representing contacted regions on the 3D object surface. 
The \textbf{frontier map $M_f$} identifies boundaries between covered and uncovered regions to guide exploration, while the \textbf{border map $M_b$} defines the valid movable range within the UV system. 
If there are holes in the target object's surface, we pre-define the corresponding regions in the UV coordinate system as covered in the coverage map. This prevents the agent from attempting to plan paths over these inaccessible areas and simplifies the learning process.

Additionally, we associate the agent with the map using an egocentric map representation \cite{chen2019learning}, which eliminates the need for the agent to learn a representation of its own pose.

\subsection{Action Space}
In real-world surface coverage tasks, continuous control offers greater flexibility and better transfer capabilities. The agent's output action is defined in the UV coordinate system as an angular velocity $\omega$. The value of $\omega$ is constrained to the range $[-45^{\circ},45^{\circ}]$. To facilitate better environmental exploration by the agent, its linear velocity is fixed as a constant. The relationship between the velocity in the UV coordinate system and the velocity of the end effector is as follows:
\begin{equation}
\begin{aligned}
    \mathbf{v}_{\text{linear}} &= \frac{\mathbf{S}(u_{t+1}, v_{t+1}) - \mathbf{S}(u_{t}, v_{t})}{\Delta t} \\
    &\quad + \frac{\mathbf{T}_{t+1} \cdot \mathbf{d} - \mathbf{T}_{t} \cdot \mathbf{d}}{\Delta t}
\end{aligned}
\end{equation}
where $d=(0,0,\delta)^T$ is the fixed offset vector between the robotic arm's control point and the contact point between the sponge and the target object.
\begin{equation}
\mathbf{\omega} = \frac{\theta}{\Delta t} \cdot \mathbf{v}
\end{equation}
Here, $\theta$ represents the rotation angle between the quaternion at the current time step $t$ and the next time step, $\mathbf{v}$ is the rotation axis, and $T_t$ denotes the rotation matrix of the end effector at time $t$. The linear velocity and angular velocity of the end effector are $\mathbf{v}_{\text{linear}}$ and $ \mathbf{\omega} $.

Assuming there are $N$ points along the path planned on the object surface, the pose of each point $p_i$ is given by $(x_i,y_i,z_i,\alpha_i,\beta_i,\gamma_i)$, where $(x_i,y_i,z_i)$ represents the position and $(\alpha_i,\beta_i,\gamma_i)$ represents the Euler angles. The values $(x_i,y_i,z_i,\alpha_i,\beta_i)$ are determined by the surface normal vector and UV coordinates, while the rotation angle of the end effector around its own Z-axis, denoted as $\gamma_i$, is set such that its rate of change equals the angular velocity $\omega$ in the UV coordinate system. This ensures that the direction of movement remains relatively perpendicular to the long edge of the contact surface.

\begin{figure*}[!t]
    \centering
    \includegraphics[width=0.75\linewidth]{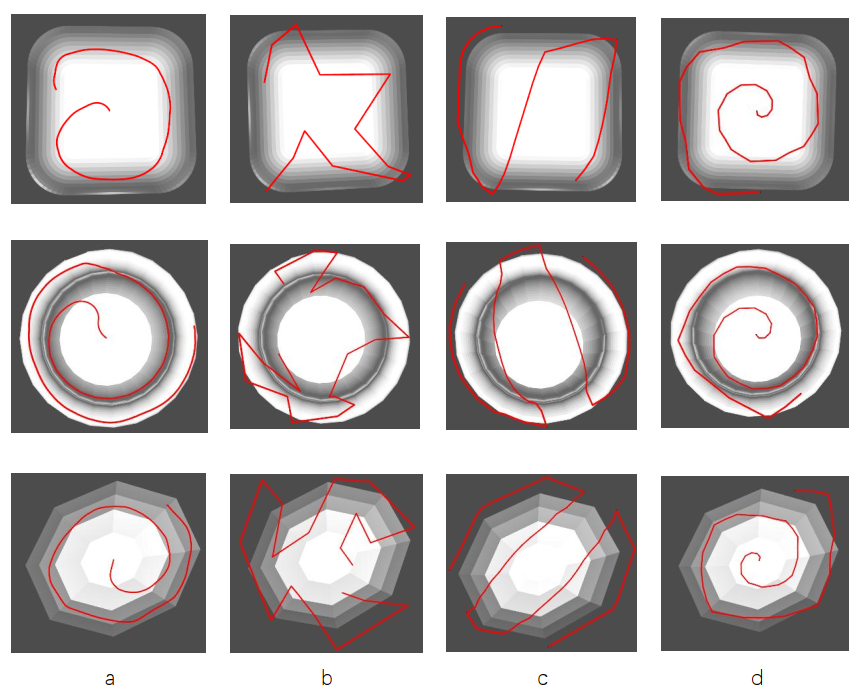}
    \caption{Paths generated by different methods on Object 1,3,4. (a) Our method, (b) SPONGE method, (c) zigzag pattern, (d) spiral pattern.}
    \label{fig:path_qulitive}
\end{figure*}

\subsection{Reward Function}
Our reward function is designed as follows:
\begin{equation}
    reward = r_c+r_{\Delta TV}+r_{const}
\end{equation}

Here, $r_c = \lambda_c \cdot N_{new}$, where $N_{new}$ denotes the number of newly covered pixels in the UV map, directly encouraging surface coverage efficiency.

Inspired by Jonnarth \textit{et al.} \cite{10.5555/3692070.3692973}, we introduce the Total Variation (TV) of the coverage map to reduce residual areas during the coverage process:

\begin{equation}
V(x) = \sum_{i,j} \sqrt{ \left(\frac{\partial x}{\partial i}\right)^2 + \left(\frac{\partial x}{\partial j}\right)^2 }
\end{equation}

In the coverage map, a high $V_C$ indicates the presence of numerous uncovered holes and streaks in the area. Reducing $V_C$ improves the agent's coverage performance.

\begin{equation}
    r_{\Delta TV}=-\lambda _{TV}(V_{C}(t)-V_{C}(t-1))
\end{equation}

To encourage the agent to explore more efficiently, a negative constant reward term $r_{const}$ is added to the reward function.

\section{Experiment}
\subsection{Implementation Details}

To validate the effectiveness of our method in 3D surface coverage tasks, we conducted comparative experiments for quantitative analysis on Objects 1–10 shown in Figure \ref{fig:objects}. We built an environment in the MuJoCo simulation platform for the task of sponge wiping on object surfaces.

Since Mujoco limits collision detection to convex geometries, it automatically replaces concave geometries with convex hulls, which prevents the detection of collisions in concave regions. To address this, we decompose complex meshes into convex hulls, thereby enabling the simulation of objects with concave features. The Mujoco simulator supports the simulation of deformable objects by treating them as composite objects consisting of spring-mass systems. The size parameters of the sponge are set consistent with those in Le \textit{et al.} \cite{le2023sponge}.

The \textbf{Coverage Area} metric is derived by mapping the covered pixels in the 2D UV map back to the 3D surface area. Specifically, we calculate the ratio of the surface area of the mesh triangles corresponding to the covered UV pixels to the total surface area of the target region.

\subsection{Comparative Methods}

The compared methods include the two-stage path planning algorithm \textbf{SPONGE} proposed by Le \textit{et al.} \cite{le2023sponge}. In its first stage, a pre-trained point cloud-based contact prediction model is used to sample path points, while the second stage computes the shortest path via a Traveling Salesman Problem (TSP) solver. Two additional baseline methods are conventional surface coverage algorithms that generate paths by sampling points based on rules in the parametric domain.

The \textbf{zigzag} method~ \cite{sheng2005tool} maps the target wiping area to a square with a side length of 2 and traverses along a zigzag path in the UV coordinate system with equivalent intervals. The path generated in the UV domain is then projected onto the object surface to form the wiping trajectory.

The \textbf{spiral} method similarly maps the object surface to a unit circle and generates a spiral wiping path originating from the center of the UV coordinate system.

We employ the Soft Actor-Critic (SAC) algorithm \cite{haarnoja2018soft} with weight sharing between the actor and critic networks and automatic temperature tuning \cite{haarnoja2018soft}. The learning rate is set to $3\cdot10^{-4}$, batch size to 256, replay buffer size to $10^6$, and discount factor $\gamma$=0.99. The action noise is configured as $3\cdot10^{-4}$, and the total training steps are set to $2\cdot10^6$. All experiments are executed on an Intel Core i9-10980XE CPU and an NVIDIA GeForce RTX 4090 GPU.

\begin{figure}[!t]
    \centering
    \includegraphics[width=0.9\linewidth]{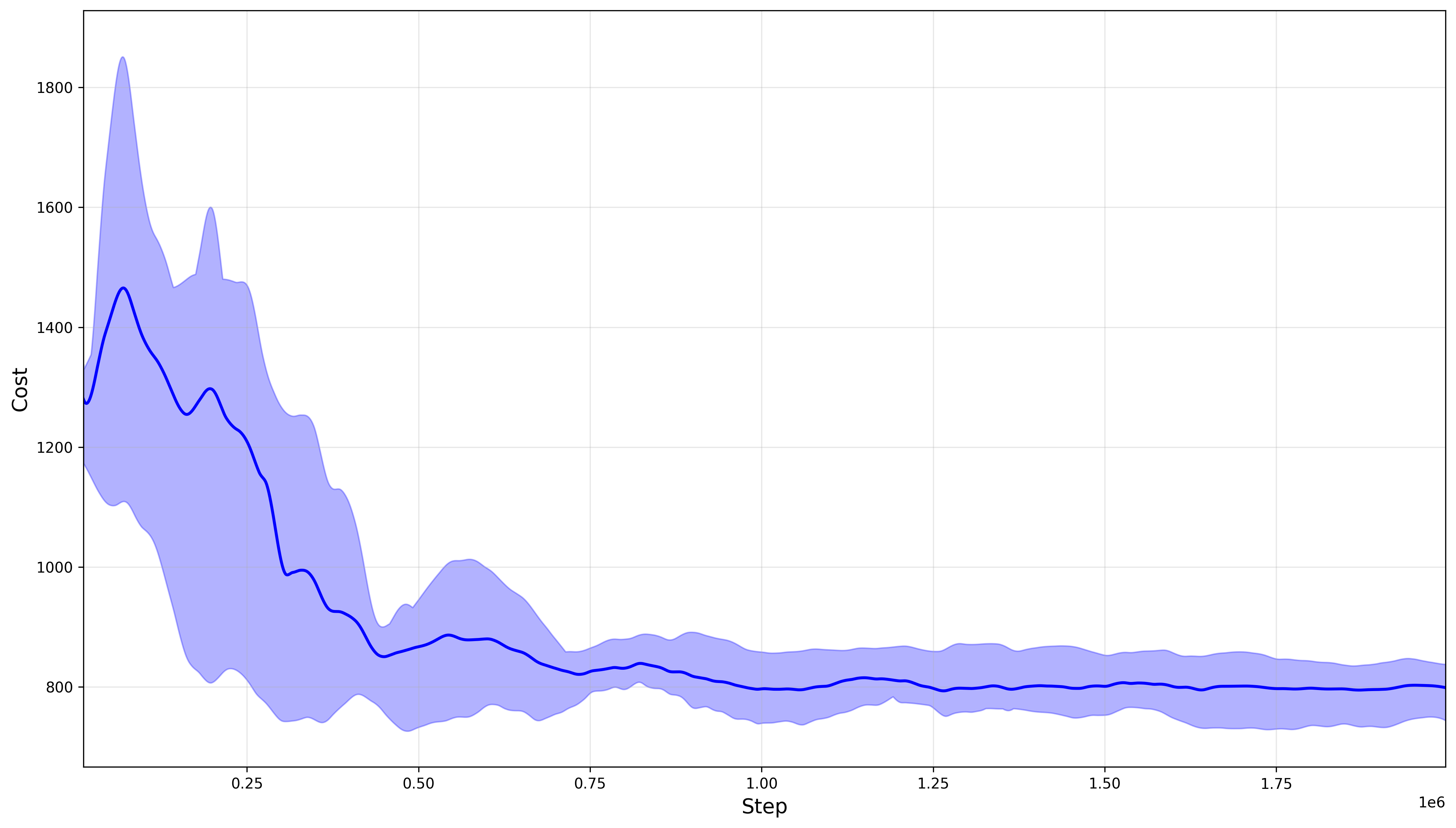}
    \caption{The convergence behavior of the agent. The y-axis represents the number of environmental steps per episode (lower is better), while the x-axis represents total training steps.}
    \label{fig:train_cost}
\end{figure}

\begin{figure}[!t]
    \centering
    \includegraphics[width=0.9\linewidth]{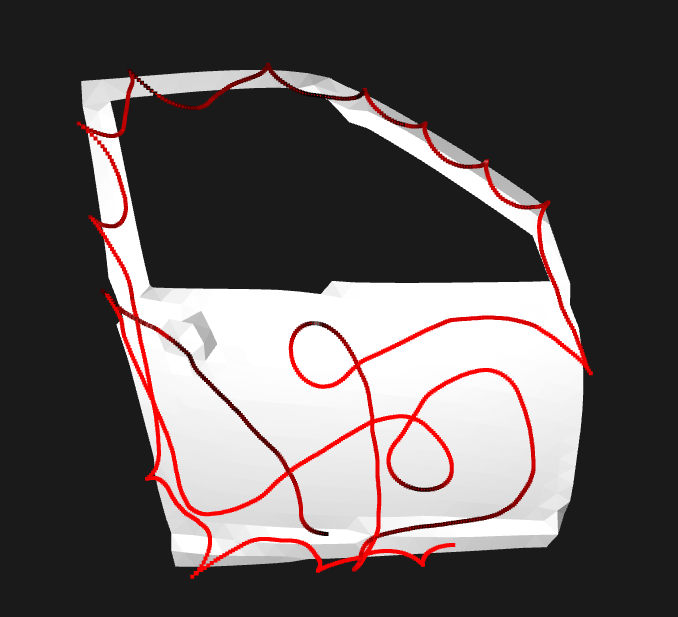}
    \caption{Our method generates paths on more complex car door surface, and the agent does not traverse through the hole regions.}
    \label{fig:car_door}
\end{figure}

\begin{figure}[!t]
    \centering
    \includegraphics[width=0.9\linewidth]{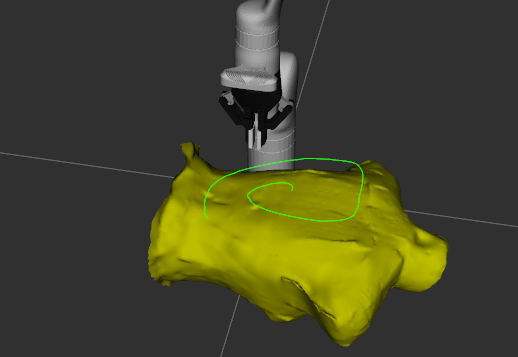}
    \caption{The yellow model in the figure represents the upper body model from the 3D reconstruction, and the green lines indicate the path trajectory of the robotic arm's end-effector.}
    \label{fig:rviz}
\end{figure}

\begin{figure}[!t]
    \centering
    \includegraphics[width=0.9\linewidth]{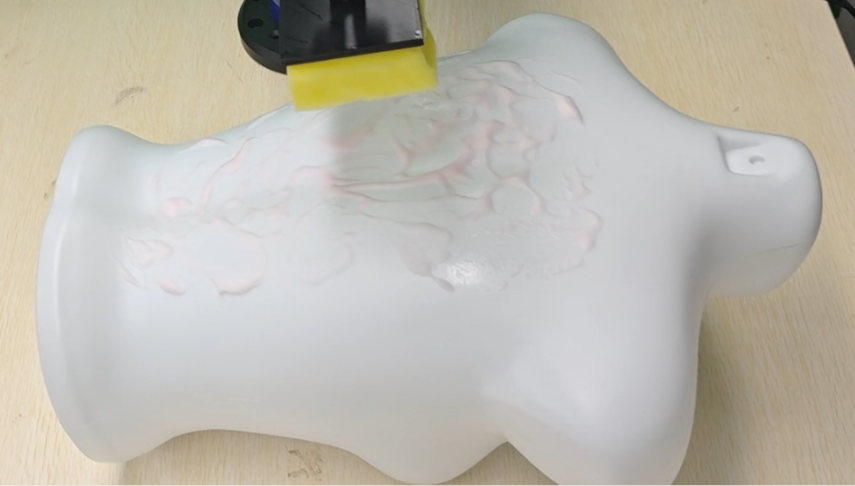}
    \includegraphics[width=0.9\linewidth]{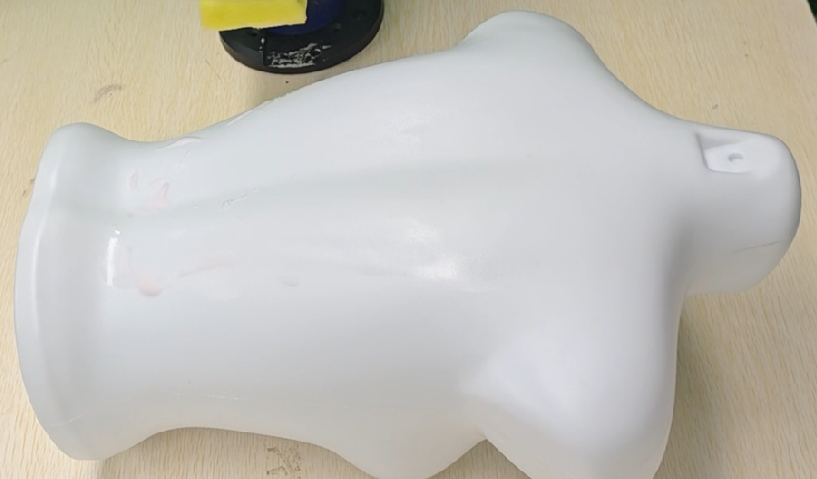}
    \caption{Qualitative experiment of wiping task on the back of a human model in the real world. The top image shows the back before wiping, with pink foam indicating the area requiring cleaning. The bottom image shows the back of the human model after cleaning.}
    \label{fig:real_experiment}
\end{figure}

\subsection{Simulation Environment Training}
In surface coverage tasks, employing reinforcement learning requires setting a target coverage ratio for the agent, otherwise, the agent may waste excessive time attempting to cover hard-to-reach or boundary regions. The episode terminates once the target coverage ratio is achieved, typically set between 90\% and 99\%. During the initial training phase in the simulation environment, the agent often requires a long path to reach the target coverage ratio. To improve training efficiency, a termination step count is configured for each object. The episode terminates when the step count reaches this predefined threshold.

Figure \ref{fig:train_cost} illustrates the convergence performance of the agent on Object 3. The vertical axis represents the number of steps required for the agent to complete an episode. It can be observed that the agent converges after $10^6$training steps. In the early stages of training, the agent's cost increases. As the agent continues to explore and trial-and-error, it eventually converges to a path with lower cost.

\subsection{Results and Evaluations}

\begin{table*}[ht]
  \centering
  \small 
  \setlength{\tabcolsep}{6pt} 
  
  \newcommand{\B}[1]{\fontseries{b}\selectfont #1}
  
  \sisetup{
    mode=text,
    detect-weight=true, 
    detect-family=true
  }
  
  \caption{Comparison of Total Path Length(m) and Coverage Area(\%) Across Different Methods on Various Containers}
  \label{tab:performance_transposed}
  
  \begin{tabular}{@{} c *{4}{S[table-format=2.2] S[table-format=3.1]} @{}} 
    \toprule
    \multirow{1}{*}{\textbf{method$\rightarrow$} } & 
    \multicolumn{2}{c}{\textbf{Ours}} & 
    \multicolumn{2}{c}{\textbf{SPONGE}} & 
    \multicolumn{2}{c}{\textbf{Zigzag}} & 
    \multicolumn{2}{c@{}}{\textbf{Spiral}} \\
    
    \cmidrule(r){2-3} \cmidrule(lr){4-5} \cmidrule(lr){6-7} \cmidrule(lr){8-9} 
    
    \textbf{Object$\downarrow$} & {Length(m)} & {Area(\%)} & {Length(m)} & {Area(\%)} & {Length(m)} & {Area(\%)} & {Length(m)} & {Area(\%)} \\
    \midrule
    
    1       & \B{0.54} & \B{97.0} & 0.81 & 89.5 & 0.68 & 94.1 & 0.84 & 93.3 \\
    2       & 0.50  & \B{98.5} & 0.81 & 97.3 & 0.57 & 98.5 & \B{0.48} & 95.8 \\
    3       & \B{1.11}  & 91.0 & 1.18 & 92.1 & 1.15 & 94.4 & 0.72 & \B{97.7} \\
    4       & \B{0.82}  & 95.0 & 1.37 & 87.6 & 1.13 & \B{98.9} & 0.91 & 96.7 \\
    5       & 0.75  & 94.0 & \B{0.71} & 94.6 & 1.06 & \B{98.7} & 0.81 & 95.4 \\
    6       & 1.12  & \B{90.0} & \B{1.06} & 88.5 & 1.33 & \B{90.0} & 1.25 & 89.7 \\
    7       & \B{0.47}  & 98.0 & 0.70 & 96.2 & 0.51 & 95.8 & 0.49 & \B{98.4} \\
    8       & \B{0.53}  & \B{99.0} & 0.55 & 98.7 & 0.60 & 98.5 & 0.54 & 98.1 \\
    9       & \B{1.06}  & 96.0 & 1.23 & \B{97.9} & 1.24 & \B{97.9} & 1.22 & 95.3 \\
    10      & \B{0.64}  & 96.0 & 1.16 & \B{100.0} & 0.85 & 91.8 & 0.67 & 92.1 \\
    \midrule
    \textbf{Total} & \B{7.54}  & 95.5 & 9.57  & 94.2 & 9.12 & \B{95.6} & 7.94 & 87.0 \\
    \bottomrule
  \end{tabular}
\end{table*}

\begin{table}[ht]
  \centering
  \caption{Cumulative sum of path rotation angle of Z-axis changes in the normal vector direction}
  \label{tab:angle_comparison}
  \begin{tabular}{@{} c S[table-format=2.2] S[table-format=2.2] @{}} 
    \toprule
    \textbf{Method$\rightarrow$} & 
    \multicolumn{1}{c}{\textbf{SPONGE}} & 
    \multicolumn{1}{c}{\textbf{Ours}} \\
    \cmidrule(r){2-3}
    \textbf{Object$\downarrow$} & 
    \multicolumn{1}{c}{$S_{|\Delta\gamma|}$} & 
    \multicolumn{1}{c}{$S_{|\Delta\gamma|}$} \\
    \midrule
    1            & 12.99              & \textbf{12.15}      \\
    2            & 21.50              & \textbf{12.72}      \\
    3            & 24.17              & \textbf{14.61}      \\
    4            & 23.04              & \textbf{15.99}      \\
    5            & \textbf{10.45}     & 14.78               \\
    6            & 17.44              & \textbf{11.10}      \\
    7            & \textbf{13.30}     & 14.82               \\
    8            & \textbf{11.29}     & 11.50               \\
    9            & 24.29              & \textbf{13.58}      \\
    10           & 29.25              & \textbf{12.57}      \\
    \midrule
    Total        & 187.73             & \textbf{133.81}     \\
    \bottomrule
  \end{tabular}
\end{table}

To evaluate the quality of the generated paths, we assessed the total path length, coverage area, and cumulative sum of rotational angle $\gamma_i$ changes for bowl-shaped objects numbered 1–10. The total path length is defined as the sum of Euclidean distances between path points, while the coverage area represents the percentage of surface patches in the target wiping region that have been contacted by the sponge. As shown in Figure \ref{fig:path_qulitive}, which illustrates the paths planned for Object 1,3,4, the gray object represents object to be wiped, and the red line denotes the planned path. Our method generates smoother paths on the target object compared to other methods. For the sponge wiping task on bowl-shaped objects 1–10, our algorithm outperforms SPONGE by nearly 27 percentage points in total path length, exceeds the traditional zigzag method by approximately 21 percentage points, and surpasses the Spiral method by about 5.3 percentage points. In terms of coverage, our method outperforms SPONGE by 1.4 percentage points, is slightly lower than the zigzag method by 0.001 percentage points, and exceeds the Spiral method by 9.8 percentage points.

For the path planned at the end-effector of the robotic arm, we also consider the variation in the Z-axis rotation angle $\gamma_i$. The cumulative sum of changes in $\gamma_i$ along the path is denoted as $S_{|\Delta\gamma|} = \sum_{i=1}^{N-1} |\Delta\gamma_i|$. In the evaluation of the cumulative change in rotational angle $\gamma_i$, our method outperforms SPONGE by 40.3 percentage points.

As shown in Figure \ref{fig:car_door}, on more complex objects such as car doors, which contain hole regions in the middle, the paths generated by our agent largely cover the door surface while avoiding the internal hole regions.

To qualitatively verify that our generated paths can accomplish the surface wiping task, we conducted experiments using a Kinova Gen3 robotic arm. The human body model in Figure \ref{fig:rviz} was obtained through relatively mature 3D reconstruction technology. The agent, trained in the MuJoCo simulation environment, outputs paths that are subsequently transferred to the real robotic arm for execution. The upper part of Figure \ref{fig:real_experiment} shows the state before wiping, with pink foam markers on the surface, while the lower part shows the state after wiping.

\subsection{Discussion}
First, although the sponge can satisfactorily complete the wiping task for the specified area in the simulation environment, in physical experiments, the end-effector should comply with the robotic arm's own joint constraints. However, the constraints of 3D surface coverage do not account for this, which leads to situations where the position and posture of the robotic arm's end-effector along the path become unreachable.

Second, the sponge in the MuJoCo simulation environment employs a spring-mass model. During its parameter configuration, a balance ought to be struck between computational time and the model's complexity. Due to the inherent limitations of the simulator, it is challenging to achieve complete consistency between the deformation of the simulated sponge and that of the real sponge. Nevertheless, with the rapid advancement of physics-based simulators for deformable objects, this gap is expected to narrow progressively.

Regarding the sim-to-real transfer without tactile sensors, we rely on the physical compliance of the deformable object (sponge) to ensure safety and contact maintenance. Although our system lacks active force feedback in the real world, the planned path includes a fixed offset $\delta$ (Eq. 5) that presses the sponge into the surface. The inherent deformability of the sponge accommodates minor geometric discrepancies between the simulation model and the real-world object, preventing damage to the rigid robot arm while ensuring effective wiping contact.

Third, current optimized path planning algorithms for surface coverage are almost offline. When dealing with movable objects, such as a patient lying in bed, the real-time capability of the algorithm ought to be considered. In the future, we may integrate a visual servoing system to enable the robotic arm to perform surface operations on moving objects.

\section{CONCLUSIONS}
We presented a novel RL-based framework for coverage path planning of deformable objects on 3D surfaces. By leveraging simulation and UV mapping with SGCNN feature extraction, our method generates efficient paths that outperform baselines in length and coverage. Real-world experiments validated the feasibility of the approach. 

However, limitations remain. First, physical joint constraints are not fully accounted for during the UV-space planning, occasionally leading to unreachable poses. Second, the sim-to-real gap in sponge deformation persists. Future work will focus on integrating real-time visual servoing to handle dynamic objects (e.g., moving patients) and considering the reachability of the generated path.
 


\bibliographystyle{IEEEtran}
\bibliography{IEEEtranBST/IEEEabrv}

\end{document}